\documentclass[10pt,journal]{IEEEtran}

\ifCLASSINFOpdf
  \usepackage[pdftex]{graphicx}
\else
\fi
\usepackage{amsmath}
\usepackage{amssymb}
\usepackage{mathrsfs}
\usepackage{color}

\usepackage{booktabs}
\usepackage{multirow}

\begin{document}
%
\title{SpVOS: Efficient Video Object Segmentation \\ with Triple Sparse Convolution}
%
%
%


\author{Weihao Lin, Tao Chen,~\IEEEmembership{Senior Member,~IEEE}, Chong Yu
\thanks{Weihao Lin and Tao Chen are with the School of Information Science and Technology, Fudan University, Shanghai 200433, China. E-mail: 21110720038@m.fudan.edu, eetchen@fudan.edu.cn. (Corresponding author: Tao Chen.)}
\thanks{Chong Yu is with Academy for Engineering and Technology, Fudan University, Shanghai 200433, China. E-mail: 21110860050@m.fudan.edu.cn}}

%
%

\markboth{IEEE TRANSACTIONS ON Image Processing,~Vol.~***, No.~***, ***~2023}%
{*** \MakeLowercase{\textit{et al.}}: SpVOS: Efficient Video Object Segmentation with Triple Sparse Convolution}
%



\maketitle

\begin{figure*}[!t]
\centering
\includegraphics[width=7.1in]{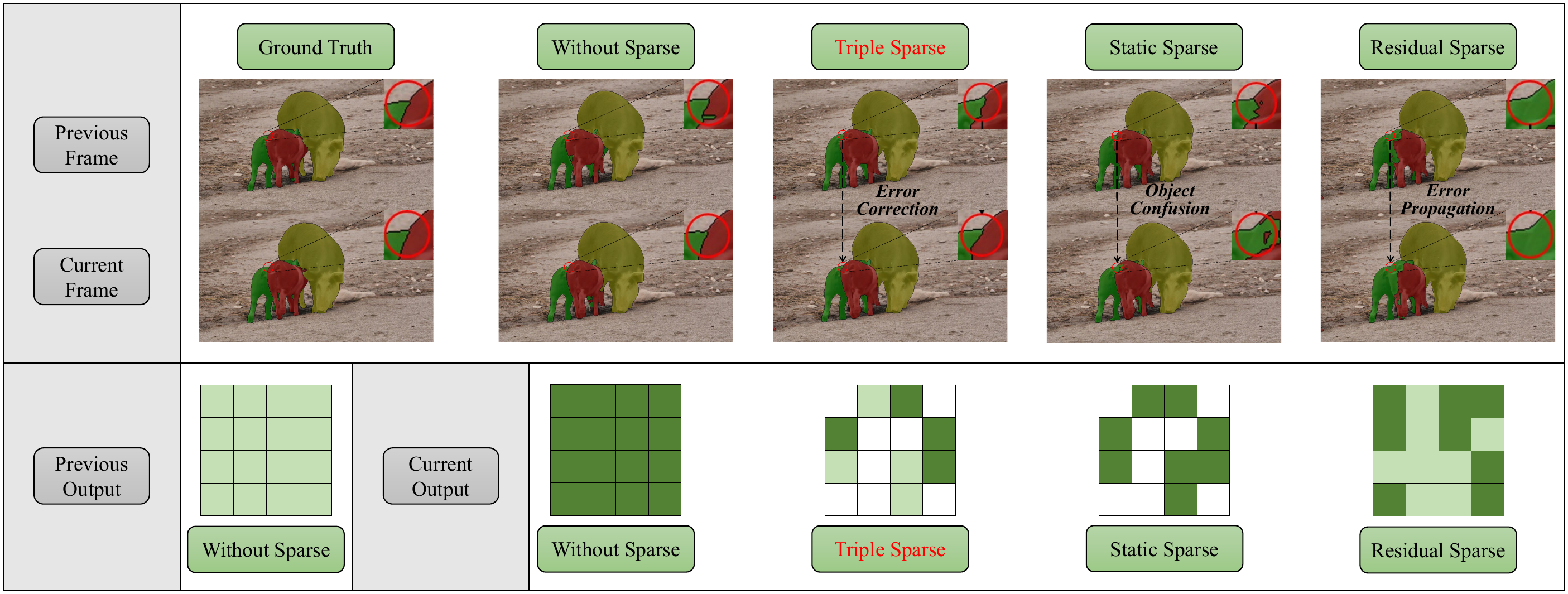}
\caption{Upper part: the segmentation results of adjacent two frames based on different sparse methods. Static sparse convolution fails to distinguish similar objects, and residual sparse convolution falls into the dilemma of error accumulation. Triple sparse convolution achieves relatively better segmentation performance, approximating the non-sparse ones. Lower part: the conceptual illustrations of different sparse methods. Each light (dark) green grid denotes a pixel of the previous (current) output, and the white grid means the value of the corresponding pixel is set to zero.}
\label{fig_1}
\end{figure*}

\begin{abstract}
Semi-supervised video object segmentation (Semi-VOS), which requires only annotating the first frame of a video to segment future frames, has received increased attention recently. Among existing Semi-VOS pipelines, the memory-matching-based one is becoming the main research stream, as it can fully utilize the temporal sequence information to obtain high-quality segmentation results. Even though this type of method has achieved promising performance, the overall framework still suffers from heavy computation overhead, mainly caused by the per-frame dense convolution operations between high-resolution feature maps and each kernel filter. Therefore, we propose a sparse baseline of VOS named SpVOS in this work, which develops a novel triple sparse convolution to reduce the computation costs of the overall VOS framework. The designed triple gate, taking full consideration of both spatial and temporal redundancy between adjacent video frames, adaptively makes a triple decision to decide how to apply the sparse convolution on each pixel to control the computation overhead of each layer, while maintaining sufficient discrimination capability to distinguish similar objects and avoid error accumulation. A mixed sparse training strategy, coupled with a designed objective considering the sparsity constraint, is also developed to balance the VOS segmentation performance and computation costs. Experiments are conducted on two mainstream VOS datasets, including DAVIS and Youtube-VOS. Results show that, the proposed SpVOS achieves superior performance over other state-of-the-art sparse methods, and even maintains comparable performance, e.g., an 83.04\% (79.29\%) overall score on the DAVIS-2017 (Youtube-VOS) validation set, with the typical non-sparse VOS baseline (82.88\% for DAVIS-2017 and 80.36\% for Youtube-VOS) while saving up to 42\% FLOPs, showing its application potential for resource-constrained scenarios.
\end{abstract}

\begin{IEEEkeywords}
Video object segmentation, convolutional neural networks, sparse convolution
\end{IEEEkeywords}

%
\IEEEpeerreviewmaketitle

\section{Introduction}
%
%
%
%

 
\IEEEPARstart{V}{ideo} object segmentation (VOS) aims to segment specific objects in videos and has a wide range of applications in real-world visual tasks, such as video editing \cite{oh2018fast}, video surveillance \cite{cohen1999detecting} and augmented reality \cite{ngan2011video}, etc. As the pixel-level semantic annotation for learning a VOS model imposes a hefty burden on humans, a lot of works study the VOS problem from the view of reducing the dependence on video annotations \cite{zhuo2019unsupervised,lu2019see,gavrilyuk2018actor,cheng2021mivos,oh2019video,yang2021collaborative}. 
Among these works, the semi-supervised VOS (Semi-VOS), where only the object labeling masks in the first frame are given to help predict the future frames, has received increasing attention recently \cite{oh2019video,yang2021collaborative}. These semi-supervised VOS methods leverage the deep convolutional neural networks (CNNs) to learn the feature correlations between adjacent frames, and have achieved promising performance.

According to the strategy of utilizing the given mask and temporal information, existing Semi-VOS pipelines can be roughly categorized into the following classes: online-learning-based  \cite{perazzi2017learning,maninis2018video}, flow-based  \cite{cheng2017segflow,luiten2018premvos}, tracking-based \cite{voigtlaender2019boltvos} and memory-matching-based \cite{voigtlaender2019feelvos,hu2018videomatch,oh2019video,seong2020kernelized}. Among them, the memory-matching-based methods \cite{yang2021collaborative,hu2021learning} tend to become the research mainstream for semi-supervised VOS, due to their competitive trade-off performance between relieving the training overhead and advancing the segmentation accuracy. However, the memory-matching-based methods are still challenged by heavy feature encoding modules like ResNet-50 and ResNet-101 \cite{he2016deep}, 
which encode the current video frame and past frames into feature memory banks. These heavy modules cause high computation costs, imposing significant computing power and resource requirements.

In this work, we focus on memory-matching-based semi-supervised VOS, and delve into sparse convolution to reduce computations of VOS in a fine-grained (layer-wise pixel-level) manner while maintaining high performance. To our knowledge, this is the first work to explore sparse convolution in VOS task. Note that some other tasks such as LiDAR BEV detection \cite{ren2018sbnet}, image classification \cite{verelst2020dynamic,dong2017more}, human pose estimation \cite{verelst2020dynamic,habibian2021skip} and video object detection \cite{habibian2021skip} have also applied the sparse convolution to do image or object-level prediction. Still, the VOS task requires pixel-level prediction for each frame, and the sparse convolution exploration in such a task encounters more challenges. For instance, a robust VOS method should deal with the complex interaction between neighboring similar objects, which may cause segmentation confusion. Further, the deformable object pose and appearance in a temporal sequence may aggravate per-frame prediction error propagation. Therefore, more effort should be put into designing and utilizing the sparse convolution to achieve efficient and robust Semi-VOS. Here we mainly discuss two critical challenges described above when exploiting the sparse convolution for VOS.

Firstly, traditional sparse convolution methods \cite{verelst2020dynamic,ren2018sbnet} mainly apply convolution kernels only on those informative sparse spatial positions that have more chances to contain an object, and set the values of other positions to zero to save the computation costs. Fig. \ref{fig_1} illustrates these sparse methods as ``Static Sparse". Although such sparse convolution can reduce spatial redundancy to some extent, it does not consider the information of previous frames and may suffer from the confusion of similar objects. For example, in ``Static Sparse" of Fig. \ref{fig_1}, the red object in the previous frame is separated from the green one. However, it is misclassified as the green object in the current frame, meaning that the object prediction consistency between adjacent frames is broken due to the lack of temporal information. In fact, the appearance of the same object in two adjacent frames usually will not change too much, and this should be considered when designing the sparse convolution for Semi-VOS task.

Secondly, even though a recent sparse convolution method considering temporal information is proposed to solve some video tasks, such as human pose estimation \cite{habibian2021skip}, it still suffers from prediction error accumulation which severely harms the performance when we apply it to VOS. In particular, this method reduces the computation amount of the current frame by adaptively reusing the information from the previous frame, which can be referred to as ``Residual Sparse" in Fig. \ref{fig_1}. It can be seen that although temporal information is utilized to maintain object prediction consistency, as a side-effect, such residual convolution inevitably brings the frame-by-frame prediction error accumulation due to the reusing of detrimental information residing in the sequence, such as the output features of the misidentified objects. For example, the red object in ``Residual Sparse" of Fig.\ref{fig_1} is misclassified as the green object in the previous frame, and this error is propagated to the current frame. Therefore, when designing the sparse convolution for VOS, detrimental information especially lasting in a temporal sequence that harms the prediction, should be identified and suppressed. 

Given the above, we propose a novel triple sparse convolution to solve the aforementioned challenges. Here, triple means there are three processing alternatives for each pixel. As illustrated in ``Triple Sparse" of Fig. \ref{fig_1}, the triple sparse convolution applies a designed triple gate on each pixel, to predict a probability for making a triple decision consisting of three policies: a skipping policy meaning the pixel's location does not require convolution operations, a reusing policy meaning the mask value of the pixel will be replaced by spatially the same one from the previous frame, and a default policy meaning the pixel will be involved into the current convolution.

Through the above design, the reused convolution features at selected positions from the previous frame, determined by the reusing choice, can ensure the object mask consistency and save the convolution re-computation overhead. As for the reused erroneous features, the triple sparse convolution can discard them by setting their values to zero based on the skipping choice, which in turn truncates the error accumulation to achieve more robust segmentation results. As shown in Fig. \ref{fig_1}, the results of ``Triple Sparse" well validate the prediction consistency of the red object. Some minor errors of the green object in the previous frame are also corrected in the current frame, thus guaranteeing a more consistent segmentation mask than the one derived from the non-sparse convolution.

To verify the efficiency and effectiveness of the proposed triple sparse convolution, we design a sparse VOS baseline, called SpVOS, which takes a video sequence as input and outputs the segmentation mask of each frame through a sparsified memory-matching-based pipeline. Note that, as the paper's central claim is to reduce the computations of VOS from a sparse convolution perspective without harming the performance. Chasing for higher accuracy to compete with state-of-the-art (SOTA) methods \cite{yang2022decoupling,cheng2022xmem}, is not our focus. The designed SpVOS merely serves as a simple and uniform baseline of sparse VOS; further accuracy improvements are possible (referred to Table \ref{table_6}) when combining the proposed triple sparse convolution with a SOTA framework, providing a future direction.

In summary, our contributions are as follows:
\begin{itemize}
    \item We reveal the high computation costs issue in existing memory-matching-based Semi-VOS works, and introduce the sparse convolution into Semi-VOS task for reducing computation costs for the first time; finally propose a sparse Semi-VOS framework called SpVOS.
    \item We deliberately design a novel triple sparse convolution layer equipped with a triple gate, which can be flexibly inserted into a typical convolutional block. Furthermore, different from traditional sparse processing, which applies sparse convolution to all samples, we develop a ``mixed sparse processing” strategy that applies sparse processing for normal frames and non-sparse processing for memory frames to optimize for a sparsity-contained objective, and achieve efficient and robust performance for memory-matching-based Semi-VOS.
    \item We conduct extensive experiments on the DAVIS \cite{perazzi2016benchmark,pont20172017} and Youtube-VOS \cite{xu2018youtube}
    benchmarks, which well validate the good efficiency-accuracy trade-off of the proposed SpVOS method. For example, on the DAVIS-2017 validation set, the proposed triple sparse convolution can reduce the overall FLOPs of all convolution layers to 42\% of the non-sparse baseline without accuracy decrease, and obtain about 5\% higher accuracy than state-of-the-art sparse convolution methods with comparable FLOPs.
\end{itemize}

This paper is organized as follows. Section \uppercase\expandafter{\romannumeral2} gives a brief review of existing VOS works, including semi-supervised VOS (Semi-VOS) and computationally efficient VOS. Several sparse convolution works for non-VOS tasks are also introduced. Section \uppercase\expandafter{\romannumeral3} presents the framework of the proposed SpVOS and details its encoder and decoder modules, and the key contributions including the triple sparse convolution with triple gate. The mixed sparse processing strategy for the SpVOS framework is also introduced. Section \uppercase\expandafter{\romannumeral4} gives the implementation details of the framework and the experimental results. Finally, a summary of this paper is concluded in Section \uppercase\expandafter{\romannumeral5}.

\section{Related Work}
\subsection{Semi-supervised Video Object Segmentation}
\subsubsection{General Methods}
Semi-supervised VOS methods generally propagate the first annotated mask to the rest frames of the given video. Early methods \cite{badrinarayanan2010label,avinash2014seamseg} mostly model the video as a graph structure and utilize an energy function to solve the optimization problem. Recently, convolutional neural networks (CNNs) have shown superior performance in semi-supervised VOS. In \cite{perazzi2017learning}, an online learning scheme is integrated with deformed object masks in CNN-based modules to adapt to specific objects in the video. Another work \cite{khoreva2017lucid} utilizes synthesized images to fine-tune the object tracker during inference. Cheng \textit{et al.} \cite{cheng2017segflow} improve the segmentation accuracy by simultaneously handling optical flow prediction and video object segmentation in an end-to-end CNN framework. Hu \textit{et al.} \cite{hu2018motion} introduce a guidance, which is generated from the last mask and current optical flow, to segment the target object accurately. What is more, specifically designed convolutional modules from other visual tasks, such as object re-identification \cite{luiten2018premvos} and tracking \cite{voigtlaender2019boltvos}, can be integrated with existing VOS methods to improve the performance.

In addition to the above works, a representative memory-matching-based method called Space-Time Memory (STM) \cite{oh2019video} appears recently, which exploits all historical frames via a memory bank, and segments specific objects in the current frame based on memorized features and a non-local \cite{wang2018non} matching, yielding significant accuracy improvement as compared with previous methods. Besides, STM does not introduce additional modules for auxiliary tasks, such as estimating optical flows \cite{cheng2017segflow,luiten2018premvos} or generating bounding boxes \cite{voigtlaender2019boltvos}. As a result, the STM structure is relatively more straightforward than other methods and has good compatibility with various matching schemes \cite{cheng2021stcn, seong2020kernelized} and memory banks \cite{liang2020video}. 

Recently, STCN \cite{cheng2021stcn} branches from STM and matches video frames without object prior, helping alleviate object confusion. AOT \cite{yang2021associating} utilizes a Transformer-based long-short-term matching module to extract object-specific features from a unified object identity embedding. Built upon STCN, XMem \cite{cheng2022xmem} introduces an Atkinson-Shiffrin memory mechanism, including sensory, working, and long-term memories, to obtain more robust results. DeAOT \cite{yang2022decoupling} decouples the propagation of video frame and object mask, further improving the performance of AOT. Although with considerable improvement, these memory-matching-based methods still suffer from high computation costs due to dense feature extractions in encoders or decoders, which will be considered and alleviated in this work.

\subsubsection{Computationally Efficient Methods}
A typical memory-matching-based VOS pipeline has four major modules: 1) a query encoder, which encodes features of the current frame; 2) a memory encoder, which extracts memories from the combination of the current frame and corresponding mask; 3) a matching module, which utilizes the attention mechanism \cite{wang2018non} to retrieve beneficial information from the memory bank; 4) a decoder, which fuses features from the query encoder with matching results to generate the predicted mask of the current frame.

In order to reduce computation costs, most existing research has primarily concentrated on diminishing the matching overhead. This has been achieved either through the selection of fewer frames to memorize, typically the first and last frames \cite{voigtlaender2019feelvos,yang2021collaborative,park2020ttvos,johnander2019generative,oh2018fast,cho2020crvos,cho2020pmvos,hong2021adaptive}, or by devising more efficient memory-matching schemes. These schemes include an adaptive memory bank \cite{liang2020video}, regional memory-matching scheme \cite{xie2021efficient}, global context \cite{li2020fast} and modified attention mechanism \cite{cheng2021mivos,seong2020kernelized,wang2019ranet}. While these approaches have effectively reduced computation costs, they have largely overlooked the potential of addressing spatiotemporal feature redundancy within the encoders and decoders. This is an area that, if explored, could offer significant reductions in computation costs. A few pioneering works have begun to venture into this territory. For instance, in \cite{park2021adaptive}, the authors innovatively reduced computation costs by pruning layers of both the query and memory encoders. This approach not only reduced the computational burden but also maintained the integrity of the encoding process, demonstrating that efficiency and effectiveness can coexist. Further expanding on this idea, the studies in \cite{wang2021swiftnet,cheng2021stcn} replaced the memory encoder with a lighter one, simultaneously adding a skip connection from the query encoder to the memory encoder. This ingenious modification reduced the computation costs and enhanced the memory encoder's representation ability, demonstrating that it is possible to reduce computation costs without compromising the quality of the encoding process. 

Moreover, out of memory-matching-based VOS, SAT \cite{chen2020state} proposes a Siamese framework and utilizes lighter feature extractors, such as Alexnet \cite{krizhevsky2017imagenet}, to reduce computations. We find that these works mainly focus on reducing the spatiotemporal redundancy in a coarse-grained manner by simply modifying partial structures of the heavy modules, leaving the most computation-intensive dense convolutions between each filter and numerous feature maps unchanged, which thus has limited computation reduction space to maintain reasonable segmentation performance. Different from these works, our work explores reducing the spatiotemporal redundancy from a new perspective of developing triple sparse convolution, which can be easily inserted into any basic convolutional block and extends the scope of optimizing the computation costs together with the segmentation accuracy. The results in the experiment section will show that our method has good flexibility to trade-off between reducing the computation costs and maintaining good segmentation accuracy.

\subsection{Coarse-to-fine Methods}
The concept of coarse-to-fine processing has been explored extensively in various computer vision fields. For example, \cite{seyedhosseini2013image} and \cite{wang2022instance} build hierarchy classifiers to fuse the predictions of different grains for better segmenting images. Ranking-based cascaded SVMs are utilized by \cite{zhang2011proposal} and \cite{zhang2015object} to assist in localizing scale-variant objects. Cascade R-CNN \cite{cai2019cascade} gradually refines the bounding box of an object based on cascaded detectors. \cite{wang2022end} proposes a cascaded framework, performing deraining and detecting in order, narrowing the gap between these two tasks. In \cite{li2017object}, a coarse-to-fine candidate region generation algorithm is proposed to capture the target objects in remote-sensing images gradually. In unsupervised VOS, the APM \cite{xi2022implicit} module processes multi-resolution features in a coarse-to-fine manner, providing details from different scales.

Different from the works above, which utilize both coarse-grained and fine-grained information in a cascaded framework, we use ``coarse” to represent the optimization of the framework is in a high level, i.e., block-wise reduction or substitution. On the other hand, we use ``fine” to denote a finer optimization, delving into layer-wise sparsity to broaden the exploration space, and reducing computations and maintaining accuracy simultaneously.

\begin{figure*}[!t]
\centering
\includegraphics[width=7.1in]{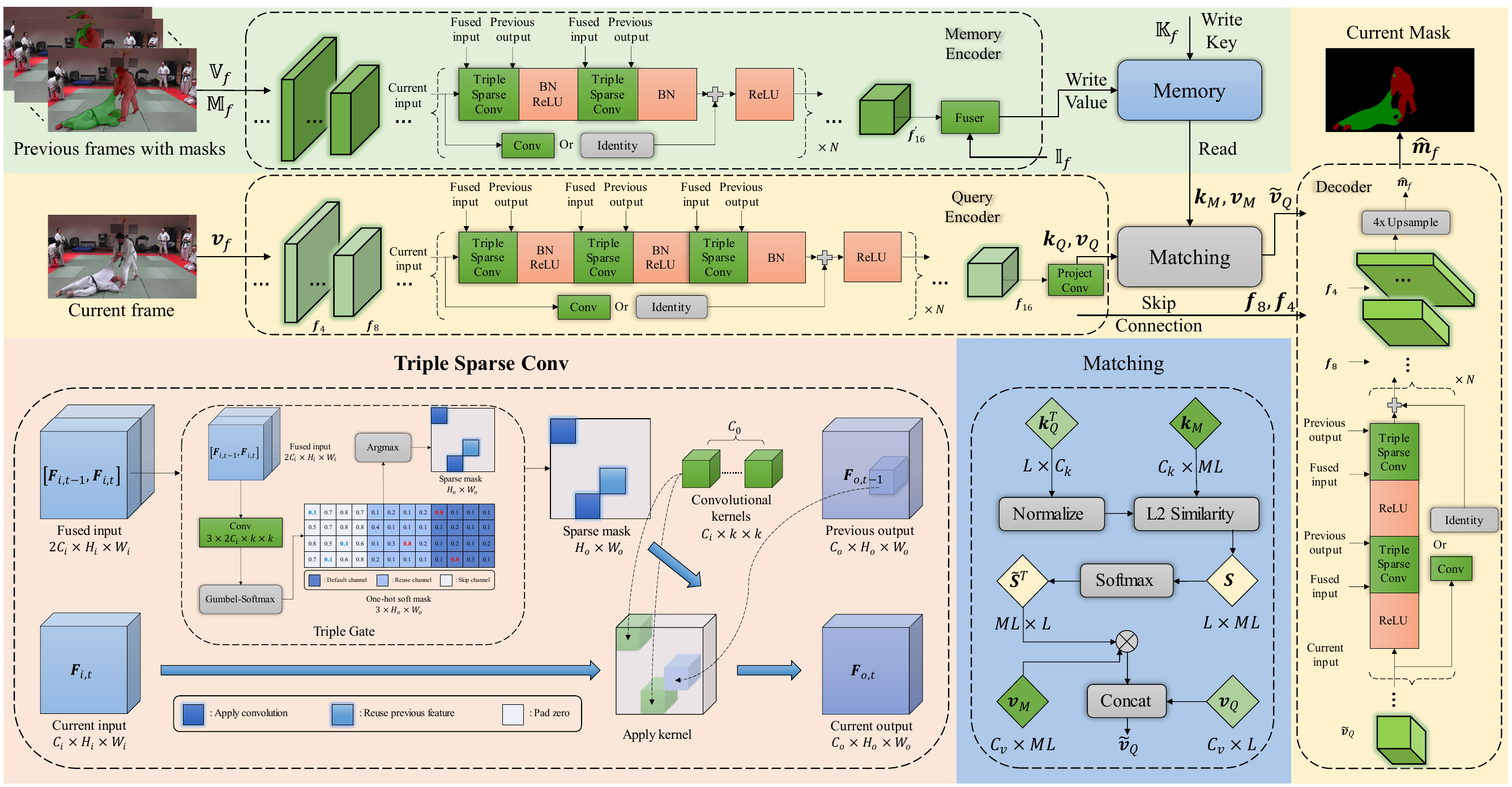}
\caption{The overall structure of the proposed SpVOS, which consists of four parts: the query encoder, the memory encoder, the matching module, and the decoder. The triple sparse convolution is proposed to replace the traditional dense convolution in each residual block of both the encoders and decoder, reducing the overall computation costs.}
\label{fig_2}
\end{figure*}

\subsection{Sparse Convolution}
Recently, one branch of lightweight CNN methods aiming at reducing the model's computation costs for image-level prediction, namely static sparse convolution \cite{verelst2020dynamic,ren2018sbnet,dong2017more}, has appeared. In general, static sparse convolution layers apply filters on partial locations of input features (or images), allowing for heavy modules to reduce spatial redundancy in a granular way. A subsequent work \cite{habibian2021skip} extends sparse convolution from the image-level task to the video-level task, by reusing features of the previous frame to reduce temporal redundancy, and is referred to as residual sparse in this work.

Although existing sparse convolution has been shown effective in LiDAR BEV detection \cite{ren2018sbnet}, image classification \cite{verelst2020dynamic,dong2017more}, human pose estimation \cite{verelst2020dynamic,habibian2021skip} and video object detection \cite{habibian2021skip}, it still faces challenges when applied in the VOS task. For example, static sparse convolution ignoring different objects' temporal feature correlations may not well distinguish similar objects in the same image. On the other hand, although the residual sparse convolution considers the temporal information, it reuses features of the previous frame, which may have poor quality in some difficult cases and will deteriorate the current frame's features and predictions. As a comparison, our triple sparse convolution exploits a designed triple gate to selectively fuse important features from the previous frame. It incorporates useful temporal clues into the sparse convolution to help identify similar objects. Those poor-quality features in the previous frame will be filtered out, by dynamically unselecting these features' pixel positions for current convolution through the triple gate.

\section{Methodology}
In this section, we first present the overall structure of the sparse VOS framework (SpVOS). Then, the major modules, including the feature encoders and decoder in the framework, are introduced, followed by a detailed discussion on the designed triple sparse convolution and the triple gate. Finally, the training objective of the whole SpVOS, and the corresponding mixed sparse VOS processing strategy are presented.

\subsection{Overall Structure}
The overall structure of the SpVOS is shown in Fig.\ref{fig_2}. Considering the input video $\mathbb{V} =\left\{ \boldsymbol{v}_1,...,\boldsymbol{v}_F \right\}$, where $\boldsymbol{v}_i\in \mathbb{R} ^{3\times H\times W}$ for $i=1,...,F$. The $F$, $H$, and $W$ are the number of frames and each frame's height and width, respectively. For the $f$-th ($f\geqslant 2$) frame $\boldsymbol{v}_f$, we define $\mathbb{V}_f$ as the set of video frames to be memorized before the $f$-th frame and $\mathbb{M}_f$ as the set of corresponding masks of these memory frames. Our task is to obtain the segmentation mask of the current frame, which is denoted as $\boldsymbol{\hat{m}}_f\in \mathbb{R} ^{H\times W}$, by utilizing the prior knowledge from $\boldsymbol{v}_f$, $\mathbb{V}_f$ and $\mathbb{M}_f$:
\begin{equation}
    \boldsymbol{\hat{m}}_f = \mathscr{N} \left( \boldsymbol{v}_f,\mathbb{V}_f,\mathbb{M}_f \right)
\end{equation}

The $\mathscr{N}$ denoting the SpVOS framework is composed of four modules, namely the query encoder, the memory encoder, the matching module, and the decoder, as illustrated in Fig.\ref{fig_2}, where the triple sparse convolution plays an important role in reducing the computation costs of both the encoders and the decoder. As a memory-matching-based framework, the temporal video sequence to be memorized is encoded into keys and values and put into the memory, referring to the memory key and the memory value. Similarly, the current query frame will be encoded into the query key and value. During the matching process, the similarity between the query key and the memory key will be calculated to modulate the memory value, to retrieve the relevant temporal information, i.e., the prior knowledge of the object-of-interest in the memory frames, with respect to (w.r.t.) the current frame. The retrieved information together with the query value and hierarchical features from the query encoder (by a skip connection), are then gradually refined in the decoder to obtain the segmentation mask. For the convenience of understanding, we only consider the single-object case in the following discussions. 

For the multi-object case, object confusion should be appropriately handled. Previously works \cite{oh2019video,yang2021associating,cheng2021stcn,cheng2022xmem,yang2022decoupling} have excessively explored this topic in non-sparse cases, generally assigning an individual memory bank for a specific object. Each memory bank provides features (concerning the specific object) from previous frames to match the current feature (of the specific object). The matching result of each object is then decoded into the segmentation mask of that object. The final mask containing all objects is obtained via soft aggregation \cite{oh2019video}. Considering sparse cases, existing sparse convolution methods \cite{verelst2020dynamic,habibian2021skip} tend to suffer more severe object confusion due to either the ignorance of temporal consistency or error accumulation. The proposed triple sparse convolution can well handle object confusion in sparse cases due to the deliberately designed triple gate according to the demonstrations of Fig. \ref{fig_1}.

\subsection{Module Design}

\subsubsection{Query Encoder}
The query encoder utilizes a typical convolutional neural network, ResNet-50 \cite{he2016deep}, to extract hierarchical features of the current frame $\boldsymbol{v}_f$ as $\boldsymbol{f}_{16}$, $\boldsymbol{f}_{8}$ and $\boldsymbol{f}_{4}$, which are $16\times$, $8\times$ and $4\times$ downsampled from $\boldsymbol{v}_f$ respectively. The $\boldsymbol{f}_{8}$ and $\boldsymbol{f}_{4}$ will be reused in the decoder through skip connection as shown in Fig. \ref{fig_2}. The $\boldsymbol{f}_{16}$ needs to be further projected into the query key and value by doing convolution with $3\times3$ kernels and $1\times1$ strides. The projected key and value are then flattened along the spatial dimensions for the subsequent matching process. We denote the flattened query key as $\boldsymbol{k}_Q\in \mathbb{R} ^{C_k\times L}$, and the flattened query value as $\boldsymbol{v}_Q\in \mathbb{R} ^{C_v\times L}$. The $C_k$ and $C_v$ are the dimensions of the feature channel with respect to (w.r.t.) $\boldsymbol{k}_Q$ and $\boldsymbol{v}_Q$. The $L$ is the flattened spatial length, i.e. $\left( H/16 \right)\times\left( W/16 \right)$.

Different from other VOS works, the convolution layers in each residual block of the query encoder (i.e., ResNet-50) in SpVOS are replaced by the designed triple sparse convolution layers. Note that the skip connection of the residual block and convolution layers in the stem of ResNet-50 (i.e., the first convolution layer) still keeps a non-sparse form, following the same settings in the image-based sparse work \cite{verelst2020dynamic}. We also try to add sparse operations to all convolution layers of ResNet-50, but the performance was not satisfactory (experimental results are shown in Table \ref{table_3}). This may be due to that too much useful information is lost by the overly-used sparse convolution in every convolution layer. 

\subsubsection{Memory Encoder}
Following \cite{cheng2021stcn,wang2021swiftnet} to reduce the memory cost, we also select a lighter neural network ResNet-18 as the memory encoder with a fusion module, denoted as ``Fuser" in Fig.\ref{fig_2}, to extract object-level temporal information. The input of the memory encoder is the concatenation of each frame to be memorized and its segmentation mask. The extracted deep feature $\boldsymbol{f}_{16}^{'}$ will be augmented, by concatenating with the previously obtained $16\times$ downsampled query feature of this memory frame. For simplicity, we denote the set of all $16\times$ downsampled query features of the memory frames as $\mathbb{I}_f$ in Fig.\ref{fig_2}. The structure of the fusion module is the same as that used in \cite{cheng2021stcn}, which consists of two residual blocks with a CBAM attention module \cite{woo2018cbam} in between. The fusion module takes the augmented feature as input and outputs the memory value that contains the object-related information of the memory frame. We reuse the query keys of the memory frames $\mathbb{K}_f$ as the memory keys, which is shown to be beneficial for keeping the consistency of the matching process as demonstrated in \cite{cheng2021stcn}.

Similar to the query encoder, we denote the memory key and value as $\boldsymbol{k}_M\in \mathbb{R} ^{C_k\times ML}$ and $\boldsymbol{v}_M\in \mathbb{R} ^{C_v\times ML}$. Note that $M$ is the number of frames to be memorized before the current frame $\boldsymbol{v}_f$. The memory key and value are the concatenations of the key and value of each memory frame, respectively. Besides, we replace the standard convolution in the residual blocks of ResNet-18 with the triple sparse convolution, to reduce the computation costs of the memory encoder.

\subsubsection{Matching}
The matching process aims to correlate the temporal information from memory frames with the current frame, to derive a new representation of the query. 

Firstly, L2 similarity, which helps to improve memory utilization and enables more robust matching as demonstrated in \cite{cheng2021stcn}, is exploited to derive the similarity matrix between $\boldsymbol{k}_Q$ and $\boldsymbol{k}_M$ as follows:
\begin{equation}
\label{eq_2}
    \boldsymbol{S}=\left( -\left\| \boldsymbol{k}_{Q}^{T} \right\|^2 +\boldsymbol{k}_{Q}^{T}\boldsymbol{k}_M \right) /\sqrt{C_k}\in \mathbb{R} ^{L\times ML}
\end{equation}
where $\| \boldsymbol{k}_{Q}^{T} \|$ means a squared sum along the feature channel, followed by an expansion operator to align with the shape of $\boldsymbol{k}_{Q}^{T}\boldsymbol{k}_M$. The $\sqrt{C_k}$ is a normalization term widely applied in similarity calculations \cite{vaswani2017attention}. 

Specifically, the Eq. \ref{eq_2} is derived from the L2 distance between $\boldsymbol{k}_Q$ and $\boldsymbol{k}_M$ along the feature dimension:
\begin{equation}
\begin{aligned}
L2\left(\boldsymbol{k}_Q,\boldsymbol{k}_M\right)&=\|\boldsymbol{k}_Q-\boldsymbol{k}_M\|^2\\
&=\left(\|\boldsymbol{k}_Q^T\|^2-2\boldsymbol{k}_Q^{T}\boldsymbol{k}_M+\|\boldsymbol{k}_M^{T}\|^2\right)\in\mathbb{R} ^{L\times ML}
\end{aligned}
\end{equation}
The $\boldsymbol{S}$ is a simplified expression with respect to the $\boldsymbol{S}_{ori}$:
\begin{equation}
    \boldsymbol{S}_{ori}=-L2\left(\boldsymbol{k}_Q,\boldsymbol{k}_M\right)/\sqrt{C_k}\in \mathbb{R} ^{L\times ML}
\end{equation}
According to \cite{cheng2021stcn}, such simplification can save computations while maintaining the final performance. The closer $\boldsymbol{k}_Q$ and $\boldsymbol{k}_M$ will yield a lower L2 distance, resulting in higher similarity. The Eq. \ref{eq_2} allows us to compute the similarity of each pair of elements ($\in\mathbb{R} ^{C_k}$) in $\boldsymbol{k}_Q$ and $\boldsymbol{k}_M$ effectively, providing a comprehensive similarity matrix.

Then the Softmax operation is applied to the similarity matrix along the dimension of memory, to obtain the normalized similarity scores as follows:
\begin{equation}
    \boldsymbol{\tilde{S}}_i=\frac{\exp \left( \boldsymbol{S}_i \right)}{\sum{\exp \left( \boldsymbol{S}_i \right)}},i=1,...,L
\end{equation}
where $\boldsymbol{S}_i$ and $\boldsymbol{\tilde{S}}_i$ are the $i$-th rows of the similarity matrix $\boldsymbol{S}$ and the normalized similarity matrix  $\boldsymbol{\tilde{S}}$ respectively. This equation is used to separately normalize each row (indexed by $i$) of the similarity matrix. The normalization aims to generate weights (that can be summed to 1) for the following weighted summation of $\boldsymbol{v}_M$.

Finally, the matching result $\boldsymbol{\tilde{v}}_Q$ is derived from the concatenation of $\boldsymbol{v}_Q$ and weighted sum of $\boldsymbol{v}_M$ based on $\boldsymbol{\tilde{S}}$:
\begin{equation}
    \boldsymbol{\tilde{v}}_Q=\left[ \boldsymbol{v}_Q,\boldsymbol{v}_M\boldsymbol{\tilde{S}}^T \right]\in \mathbb{R} ^{2C_v\times L}
\end{equation}
This equation fuses the current feature value with the weighted memory values, generating a refined representation. The weighting is based on the normalized similarity, which can emphasize the more relevant features in the memory, thereby enhancing the matching result.

\subsubsection{Decoder}
By skip-connecting the extracted query features $\boldsymbol{f}_{8}$ and $\boldsymbol{f}_{4}$ to the decoder, the matching result $\boldsymbol{\tilde{v}}_Q$ is gradually refined to obtain the final segmentation mask $\boldsymbol{\hat{m}}_f$. The decoder structure follows a general one used in \cite{oh2019video}, which consists of several residual blocks and upsampling operators. The details of the decoder will not be discussed in this work and can be referred to \cite{oh2019video}. Similar to the encoder modules, the convolution layers in the residual blocks of the decoder also employ the triple sparse convolution.

\subsection{Triple Sparse Convolution}

\subsubsection{Workflow}
A detailed workflow of the proposed triple sparse convolution is illustrated at the bottom left of Fig. \ref{fig_2}. For an inserted triple sparse convolution layer in an encoder (or decoder), its inputs for the current frame and previous frame are denoted as $\boldsymbol{F}_{i,t}$ and $\boldsymbol{F}_{i,t-1}$ respectively, both of which belong to $\mathbb{R} ^{C_i\times H_i\times W_i}$. Based on the concatenation of $\boldsymbol{F}_{i,t}$ and $\boldsymbol{F}_{i,t-1}$, a mask for determining how to apply the triple sparse convolution on a selected position at the current frame will be generated by a designed gate function, which is called the triple gate to be introduced in the following section. The mask $M$ is computed as follows:
\begin{equation}
    \boldsymbol{M}_s = \mathscr{G}\left(\left[ \boldsymbol{F}_{i,t-1},\boldsymbol{F}_{i,t} \right] \right),\boldsymbol{M}_s\in\left\{ 0,1,2 \right\} ^{H_o\times W_o}
\end{equation}
where $\mathscr{G}$ denotes the triple gate function, $\left[ *,* \right]$ means the concatenation operation along the channel dimension and $H_o$, $W_o$ are the output size. The $\boldsymbol{M}_s$ is the triple sparse mask, where triple means the values of the sparse mask are three-fold, namely $0$, $1$, and $2$ for each spatial location. Different values in the sparse mask correspond to different policies. Specifically, $0$ means the skipping policy and spatial locations with this policy will be filled with zeros; $1$ means the reusing policy, and values of the previous input will be directly copied to the corresponding spatial locations of the current input without re-computation; $2$ means the default policy and sparse convolution kernels will be applied to spatial locations with this policy.

The above three policies can be formulated as follows:
\begin{equation}
    \boldsymbol{F}_{o,t} = \mathbb{I}\left[ \boldsymbol{M}_s=2 \right]\cdot\boldsymbol{F}_{o,t}^{'}+\mathbb{I}\left[ \boldsymbol{M}_s=1 \right]\cdot\boldsymbol{F}_{o,t-1}
\end{equation}
where $\boldsymbol{F}_{o,t}$ and $\boldsymbol{F}_{o,t-1}$ are the current output and previous output of the triple sparse convolution layer, respectively. The $\boldsymbol{F}_{o,t}^{'}$ is the non-sparse output derived from applying normal convolution kernels on each spatial location, and $\mathbb{I}\left[* \right]$ is the indicator function. It is worth mentioning that the above formula is from the theoretical view to elaborate the triple sparse convolution, where it can be observed that the default policy introduces the 
highest convolution computation costs. To speed up this process, we implement a specific CUDA kernel for the triple sparse convolution, to significantly reduce the latency of dense convolutions.

\subsubsection{Triple Gate}
The implementation of the triple gate $\mathscr{G}$ is illustrated in the dashed box of the triple sparse convolution layer in Fig. 2. A standard convolution layer is first applied on the fused input of the triple gate, to extract the original probability of applying each policy on the current feature map. After that, we employ the Gumbel-Softmax technique to perform probabilistic sampling based on the probability generated by the convolution, which intuitively introduces randomness for exploring more diverse sparse patterns, and turns out to be advantageous over simply applying Softmax to achieve a better trade-off between sparsity and accuracy \cite{xie2020spatially}. Specially, given the unnormalized original probability $\boldsymbol{P}\in\mathbb{R} ^{3\times H_o\times W_o}$, the Gumbel-Softmax technique \cite{jang2016categorical} acts as follows:
\begin{equation}
\label{eq_9}
\begin{aligned}
    \boldsymbol{\tilde{M}}_s&=\text{Softmax} \left( \frac{\log \left( \text{Softmax} \left( \boldsymbol{P} \right) \right) +\boldsymbol{G}}{\tau} \right) \\
    \boldsymbol{G}&=-\log \left( -\log \left( \boldsymbol{U} \right) \right),\boldsymbol{G} \text{ and } \boldsymbol{U}\in \mathbb{R} ^{3\times H_o\times W_o}
\end{aligned}
\end{equation}
where $\boldsymbol{G}$ is the Gumbel noise generated from $\boldsymbol{U}$, and each element of $\boldsymbol{U}$ is independent and identically distributed (i.i.d.) and sampled from a uniform distribution ranging from $0$ to $1$. The $\tau$ is a temperature factor set to $1$ unless specified otherwise. Both Softmax operations in the above formula are performed along the channel dimension. The $\boldsymbol{\tilde{M}}_s$ is the soft sparse mask, and the sparse mask $\boldsymbol{M}_s$ can be obtained by applying Argmax operation to the soft sparse mask along the channel dimension.

Note that the Argmax operator is non-differentiable, which hinders the back-propagation gradients. In order to solve this problem, we resort to the straight-through estimator as in \cite{bengio2013estimating,verelst2020dynamic}. Specifically, the non-differentiable sparse mask $\boldsymbol{M}_s$ is used for the sparse convolution only in the forward pass, and the differentiable soft sparse mask $\boldsymbol{\tilde{M}}_s$ is used for calculating the gradients in back-propagation instead.

\subsection{Training Objective}

The training objective of SpVOS $\mathcal{L}$ consists of two components, including the segmentation loss $\mathcal{L}_{seg}$ and the sparse loss $\mathcal{L}_{sp}$, which is formulated as:
\begin{equation}
\mathcal{L} = \mathcal{L}_{seg} + \gamma\mathcal{L}_{sp}
\end{equation}

Following \cite{cheng2021stcn}, the $\mathcal{L}_{seg}$ is the bootstrapped cross entropy loss between the predicted mask $\boldsymbol{\hat{m}}_f$ and the ground truth $\boldsymbol{m}_f$. The $\mathcal{L}_{sp}$ comprises a layer-wise and global sparse loss. Given a desired sparse target, the $\mathcal{L}_{sp}$ firstly restricts the sparsity of each layer to meet the sparse target, then gradually relaxes the layer-wise sparse target to encourage more diversified sparse patterns while keeping the global sparse target satisfied. As the feature redundancy for different layers may vary from each other, diversified sparse patterns can fully exploit the sparsity of each layer, and try to reach an overall optimal performance under a fixed global sparse target.

Especially, suppose there are $K$ triple sparse convolution layers in total, and the sparse target is set to $t_s$. Then for the $k$-th layer, its layer-wise sparse loss is defined as:
\begin{equation}
\begin{aligned}
    \mathcal{L}_{layer,k}=\left( \max \left( s_k-t_{upper},0 \right) \right) ^2 \\
    + \left( \max \left( t_{lower}-s_k,0 \right) \right) ^2
\end{aligned}
\end{equation}
where $t_{upper}$ and $t_{lower}$ are the upper bound and the lower bound of the relaxed sparse target, which are both initialized to $t_s$ at the beginning of the training and then gradually relaxed to $1$ and $0$ respectively. We utilize the same relaxing method as \cite{verelst2020dynamic}, and more details about $t_{upper}$ and $t_{lower}$ can be referred to in \cite{verelst2020dynamic}. The $s_k$ is the layer-wise sparsity of the $k$-th layer, determined by the default part only, i.e. $\boldsymbol{M}_{d,k}=\mathbb{I}\left[ \boldsymbol{M}_{s,k}=2 \right]$, w.r.t. the corresponding sparse mask $\boldsymbol{M}_{s,k}$ of this layer. Formally, we have $s_k=\left( \sum{\boldsymbol{M}_{d,k}} \right) /\left( H_{o,k}W_{o,k} \right)$, and $H_{o,k}$, $W_{o,k}$ are the height and the width of the $k$-th sparse mask respectively. Then, the layer-wise sparse loss of all layers can be formulated as $\mathcal{L}_{layer}=\sum_{k} \mathcal{L}_{layer,k}$.

In order to obtain the global sparse loss, we utilize the 
pixel-level FLOPs of the triple sparse convolution layer to 
indicate the computation costs of different layers. For the $k$-th layer, we denote its pixel-level FLOPs as $c_k$, which is equal to the element number of the convolution kernels, i.e., $k_k\times k_k\times C_{i,k} \times C_{o,k}$. Then, the global sparse loss $\mathcal{L}_{global}$ can be expressed as follows:
\begin{equation}
    \mathcal{L}_{global}=\left( \frac{\sum_k{\left( c_k\left( \sum{\boldsymbol{M}_{d,k}} \right) \right)}}{\sum_k{\left( c_kH_{o,k}W_{o,k} \right)}}-t_s \right) ^2
\end{equation}

Finally, the sparse loss $\mathcal{L}_{sp}$ is the summation of the layer-wise sparse loss and the global sparse loss, denoted as follows: 
\begin{equation}
\mathcal{L}_{sp}=\mathcal{L}_{layer}+\beta\mathcal{L}_{global}.
\end{equation}

\begin{table*}[!t]
\caption{The quantitative evaluation results of different sparse methods on DAVIS-2017 and Youtube-VOS 2018 validation set. We also report the comparison of the parameter size (Params) and the number of floating-point operations (FLOPs) as the metric of computation costs, which is derived from averaging the FLOPs of processing all frames on the DAVIS-2017 validation set. Frames per second (FPS) is used to measure the practical runtime of methods.
. For ``Triple", ``w/o op." and ``w/op." mean implementation without/with a suitable optimization. The optimal and sub-optimal values of each column are marked in bold and italics, respectively.}\label{table_1}
\begin{center}
\setlength{\tabcolsep}{7.5pt}
\renewcommand{\arraystretch}{1.2}
\begin{tabular}{cccccccccccc}
\toprule[0.5pt]
\multirow{2}{*}{Methods} & \multicolumn{3}{c}{DAVIS-2017}                             & \multicolumn{5}{c}{Youtube-VOS}                                                 & \multirow{2}{*}{Params} & \multirow{2}{*}{FLOPs} & \multirow{2}{*}{FPS} \\ 
\cmidrule[0.5pt](r){2-4}\cmidrule[0.5pt](r){5-9}
                         & $\mathcal{J}\&\mathcal{F}$ & $\mathcal{J}$ & $\mathcal{F}$ & Overall & $\mathcal{J}_S$ & $\mathcal{F}_S$ & $\mathcal{J}_U$ & $\mathcal{F}_U$ &                         &     &                   \\ \midrule[0.5pt]
Static                   & 77.66\%                      & 74.66\%         & 80.67\%         & 76.39\%   & 76.92\%           & 81.51\%           & 69.64\%           & 77.48\%           & \textit{208.00M}                 & \textit{126.24G}    & 16.74            \\
Residual                 & 78.08\%                      & 75.23\%         & 80.93\%         & 74.11\%   & 75.73\%           & 79.88\%           & 66.58\%           & 74.25\%           & \textit{208.00M}                 & \textbf{125.23G}       & 15.78        \\
Non-sparse               & \textit{82.88\%}                      & \textit{79.59\%}         & \textit{86.18\%}         & \textbf{80.36\%}   & \textbf{80.65\%}           & \textbf{85.21\%}           & \textbf{74.13\%}           & \textbf{81.46\%}           & \textbf{207.78M}                 & 220.62G  & \textit{18.67}              \\ \midrule[0.5pt]
Triple (w/o op.)                & \textbf{83.04\%}                      & \textbf{79.83\%}         & \textbf{86.25\%}         & \textit{79.29\%}   & \textit{79.01\%}           & \textit{83.55\%}           & \textit{73.29\%}           & \textit{81.33\%}           & 209.06M                 & 128.42G  & 15.38 \\
Triple (w/ op.)                & \textbf{83.04\%}                      & \textbf{79.83\%}         & \textbf{86.25\%}         & \textit{79.29\%}   & \textit{79.01\%}           & \textit{83.55\%}           & \textit{73.29\%}           & \textit{81.33\%}           & 209.06M                 & 128.42G  & \textbf{24.32} \\ \bottomrule[0.5pt]
\end{tabular}
\end{center}
\end{table*}

\subsection{Mixed sparse processing}
Traditional sparse processing \cite{dong2017more,ren2018sbnet,verelst2020dynamic,habibian2021skip} applies sparse convolution to all samples, which might be suitable for visual tasks like image classification and human pose estimation. However, for the memory-matching-based Semi-VOS, applying sparse processing for all frames without differentiation will greatly degrade the performance, as some frequently reused memorized features, such as features of the first frame with accurate annotations, play very important roles in the matching process, and thus non-sparse processing may be required for these important frames to keep as much useful features as possible.

In order to relieve this problem, we adopt a mixed sparse processing strategy which incorporates the sparse and the non-sparse processing together, by only applying sparse convolution to the frames that do not require memorizing to maintain high performance as non-sparse processing does, 
meanwhile reducing the computation costs to that of the sparse processing but with slightly more computation of the memory frames. The following section will discuss more details about mixed sparse processing during training and inference.

\section{Experiments}

In this section, we first introduce the Semi-VOS task's mainstream datasets and experimental settings. Then, we compare the existing sparse methods with our proposed method for Semi-VOS, in terms of segmentation accuracy and computation costs. After that, we conduct extensive ablation studies, including insight analyses and qualitative visualizations, to evaluate the effectiveness of the proposed method.

\subsection{Datasets}

\subsubsection{DAVIS-2017}
The DAVIS 2017 validation set is a multi-object segmentation benchmark and consists of $59$ objects in $30$ videos. For the performance on DAVIS 2017 validation set, we utilize the official evaluation code \cite{perazzi2016benchmark} to measure the region similarity $\mathcal{J}$, the contour accuracy $\mathcal{F}$, and the average of these two metrics, i.e., $\mathcal{J}\&\mathcal{F}$.

\subsubsection{Youtube-VOS}
The Youtube-VOS validation set is much larger than the DAVIS 2017 validation set. It consists of $507$ videos with $65$ seen object categories from the training set and additional $26$ unseen categories in the validation set. For the performance on the Youtube-VOS validation set, we upload our results to the online evaluation server \cite{xu2018youtube}, which separately evaluates the region similarity and the contour accuracy on seen object categories ($\mathcal{J}_S$ and $\mathcal{F}_S$) and unseen object categories ($\mathcal{J}_U$ and $\mathcal{F}_U$). The average of these four metrics will be calculated and used as the overall evaluation criteria. 

\subsection{Settings}

\subsubsection{Training and Testing Details}

We utilize the widely used VOS datasets, including DAVIS \cite{perazzi2016benchmark,pont20172017} and Youtube-VOS \cite{xu2018youtube} to train the model. Following \cite{oh2019video}, for a given video in the dataset; three frames are sampled to form the training video sequence in a curriculum learning manner \cite{yang2015weakly}. We follow the same configuration in \cite{cheng2021stcn} to perform random affine, random resized crop, random horizontal flip, random grayscale, and color glitter to augment the training sequence.

During training, we set the first two frames as the memory frames. The annotated first frame is used to initialize the memory bank. The mask of the second frame is predicted based on the memory extracted from the first frame, and then utilized to generate the memory of the second frame. Finally, the segmentation result of the third frame is obtained based on the memory of the previous two frames. During inference, the memory frames are selected after every five/four frames in addition to the annotated frames on DAVIS/Youtube-VOS, respectively. 

For the sparse implementation, previous methods generally process all frames with sparse convolution during both the training and inference stages. However, as mentioned before, the sparse processing of the memory frames will greatly degrade the performance. Therefore, we propose mixed sparse processing for both training and inference. The mixed sparse training only applies sparse convolution to the third frame while processing the first two memory frames in the non-sparse form, i.e., we do not utilize the triple gate to generate the sparse mask and merely apply typical convolution with the same filter as the sparse one. Similarly, for the mixed sparse inference, all memory frames, including the annotated frames, are processed in the non-sparse form. 

\subsubsection{Implementation Details}

We adopt the AdamW \cite{loshchilov2017decoupled} optimizer with a weight decay of $0.05$ to train the model. The batch size for training is set to $16$ using 4 RTX 3090 GPUs. The number of training iterations is set to $160000$, and a linear learning rate warm-up from $1e-6$ to $2e-5$ is applied for the first $8000$ iterations, followed by a cosine annealing schedule starting from $2e-5$ at the $32000$-th iteration and gradually decaying to $1e-6$ at the end of training. The hyperparameters $\gamma$ and $\beta$ are both set to $1$. The $C_k$, $C_v$ are set to $64$ and $512$ respectively. We use PyTorch 1.8.2 \cite{paszke2017automatic} to implement our method.

\begin{figure*}[!t]
\centering
\includegraphics[width=7.1in]{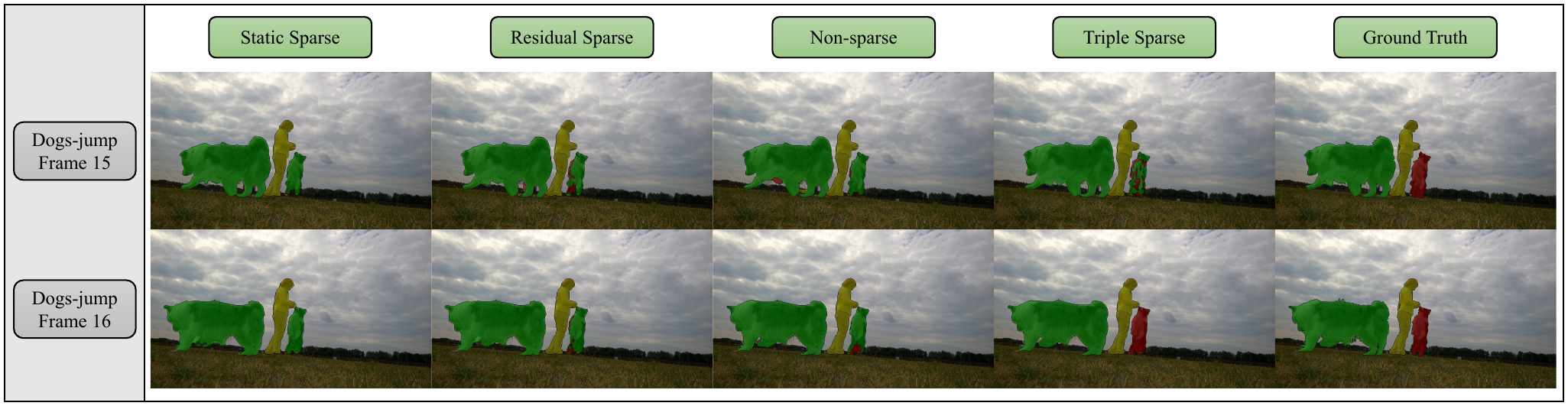}
\caption{The qualitative results of different sparse methods on the DAVIS-2017 validation set. In the ``Dogs-jump Frame 16" image, the proposed triple sparse method can self-correct the misidentified red dog in the previous frame while other sparse methods and even the non-sparse method cannot handle it well.}
\label{fig_davis}
\end{figure*}

\subsubsection{Baselines} 

To the best of our knowledge, no precedent applies sparse convolution to the Semi-VOS task so far. In order to compare the proposed triple sparse convolution with other sparse methods for Semi-VOS, we attempt to tune several existing sparse methods developed for other vision tasks for the Semi-VOS task, and use them as the sparse baselines in this work. The following will introduce the details about our designed Semi-VOS sparse baselines, and a non-sparse baseline is also given to show the superiority of sparse-based methods. 

\begin{figure*}[!t]
\centering
\includegraphics[width=7.1in]{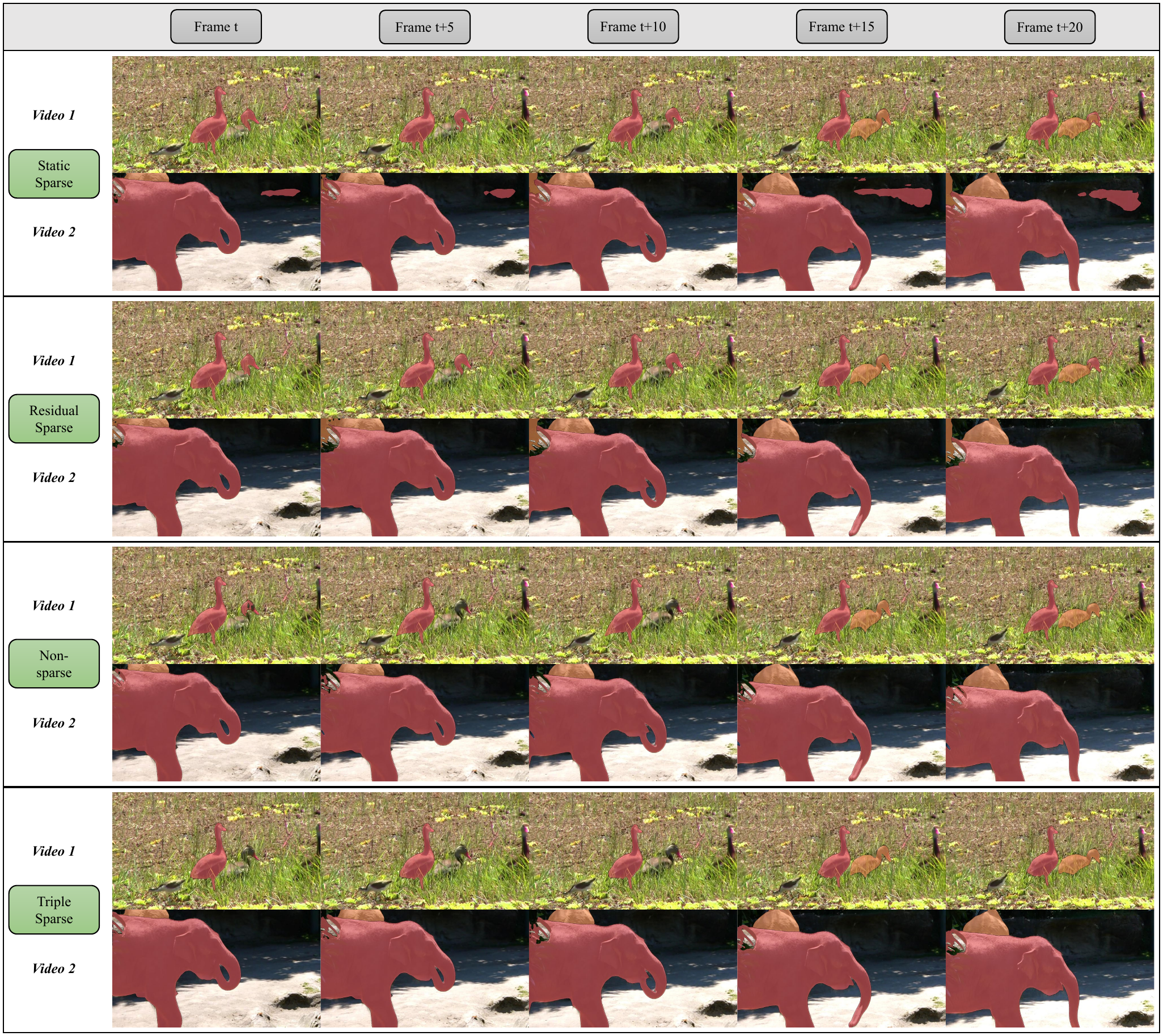}
\caption{The qualitative results of different sparse methods on Youtube-VOS validation set. The proposed triple sparse method can achieve temporally consistent results in both ``Video 1" and ``Video 2", and even perform better than the non-sparse method in the $t$-th frame of ``Video 1" when identifying similar objects. The static sparse method cannot well distinguish similar objects in ``Video 1" and suffers from severe background disturbance in ``Video 2". The residual sparse method also suffers from object confusion, such as the misidentified objects in the right part of ``Video 1" and the upper-left part of ``Video 2" respectively.}
\label{fig_youtube}
\end{figure*}

\textbf{Static Sparse}: The static sparse convolution \cite{ren2018sbnet,verelst2020dynamic,dong2017more} is designed for reducing the computation costs of image-level tasks, by adaptively applying convolution kernels to partial locations of the current input without using any temporal information. For the static sparse baseline, we replace each triple sparse convolution layer in SpVOS with the sparse method in \cite{verelst2020dynamic}. Other configurations of the static sparse baseline keep the same as the triple sparse ones mentioned before.

\textbf{Residual Sparse}: The residual sparse convolution \cite{habibian2021skip} is designed for video-level tasks, which selectively retrieves some output pixel values of the previous frame to reduce the computation of the current frame. Similar to the static sparse baseline, we substitute all triple sparse convolution layers with the residual sparse convolution layers in \cite{habibian2021skip}. 

The training and inference configurations are the same as those of triple sparse models, but with a different training objective as in \cite{habibian2021skip}, which does not consider the sparse target but simply uses the global computation costs as the loss. We experimentally find that the sparsity metric of residual sparse models always converges to a fixed value (approximately $0.1$) no matter how the sparse target changes, and the training is found to be more stable without the sparse target. Therefore, we remove the sparse target for residual sparse models to ensure both reliable and stable results.  

Moreover, to achieve a fair comparison, we set the sparse target for both static sparse and triple sparse models to be $0.1$ during training, to achieve comparable computation costs with the residual sparse models. The two models are found to finally converge with approximately $0.1$ sparsity.

\textbf{Non-sparse}: For the non-sparse baseline, we just utilize the traditional convolution layers to replace the sparse layers in each residual block of SpVOS, and obtain a typical memory-matching-based Semi-VOS framework. As there is no sparse operation involved, the sparsity loss is removed from the training objective.  Other training and inference configurations keep unchanged.

\subsection{Results}

\subsubsection{Results on DAVIS-2017}

The proposed model is firstly compared with the aforementioned sparse baselines and the non-sparse baseline on the DAVIS-2017 validation benchmark. Results in the ``DAVIS-2017" part of Table \ref{table_1} show that our proposed triple sparse method exceeds both static sparse and residual sparse methods by a large margin of around 5\%. One possible reason is that our method attempts to reduce the computation costs by exploring spatiotemporal redundancy with the help of the proposed triple gate, which can adaptively reuse informative features and discard inconsistent features from the previous frame to alleviate both object confusion and error accumulation.

As shown in Fig. \ref{fig_davis}, the proposed triple sparse method can correct some erroneous segments to obtain more accurate results, such as the red dog in ``Dogs-jump". However, the static sparse method suffers from object confusion in ``Dogs-jump", which cannot be corrected due to the lack of temporal guidance. The residual sparse method tends to worsen the erroneous results in ``Dogs-jump", as it cannot filter out the detrimental features from the previous frame. Noteworthily, the proposed triple sparse method can even achieve better segmentation results than the non-sparse method for ``Dogs-jump", probably because the method benefits from the adaptive reusing and skipping policies of the triple gate, which renders the model more capability to self-correct the erroneous segments at a larger scale\footnote{The non-sparse method, or to say the memory-matching-based method, can also self-correct the errors to a certain extent, due to the utilization of memorized features, but such ability still suffers from error accumulation for large-scale errors as can be seen from ``Dogs-jump" in Fig. \ref{fig_davis}.}.

\subsubsection{Results on Youtube-VOS}

We further compare the performance of different methods on the more challenging Youtube-VOS dataset. The ``Youtube-VOS" part of Table \ref{table_1} summarizes the results of competing methods on the Youtube-VOS validation set. Although a slight performance drop of about 1\% can be observed as compared with the non-sparse method, the proposed method can still achieve the top $\mathcal{J}\&\mathcal{F}$ score of 79.29\% among all sparse variants, outperforming the static sparse method and the residual sparse method by a large margin of approximately 3\% and 5\%, respectively.

Fig.\ref{fig_youtube} illustrates the qualitative results of the Youtube-VOS validation set. Similar to the observation on DAVIS-2017, the triple sparse method can well handle object confusion and error accumulation. In contrast, the other two sparse methods cannot obtain consistent results as the non-sparse method, due to either the distractions from the background and similar objects or the accumulation of errors.

\subsubsection{Complexity Comparison}

To study the complexity and cost of the sparse methods, we report the parameter size (Params), floating-point operations (FLOPs), and frames per second (FPS) on a V100 GPU. As shown in the rightmost part of Table \ref{table_1}, the proposed triple sparse method introduces a much smaller computation overhead (about 42\% reduction w.r.t. the non-sparse method) and a negligible increase in the parameter size (about 0.6\% more than the non-sparse method). When compared with other sparse methods (static and residual sparse), the triple sparse method has a slightly higher computation overhead with additional 1.06M (about 0.5\%) parameters and 3.19G (about 2.5\%) more FLOPs than the residual sparse method, and 2.18G (about 1.7\%) more FLOPs than the static sparse method. Such a slight computation increase is acceptable w.r.t. the significantly higher performance achieved as mentioned before. For runtime, sparse methods without suitable engineering optimization, such as utilizing a specific CUDA kernel, generally suffer from inferior FPS compared to the non-sparse baseline. Note that, the FPS of the proposed triple sparse convolution without a suitable optimization does not differ much from the static and residual ones (only about 1.4 lower than the static one and 0.4 lower than the residual one). With the designed CUDA kernel, the proposed triple sparse convolution can be much faster than the other two sparse methods and renders the baseline $1.3\times$ runtime speed up, showing the potential of practical applications.

\subsection{Ablation Studies}

In this section, ablative experiments are conducted to study the effectiveness of different sparse processing strategies and model structures with the effect of the temperature in Gumble-Softmax. We also visualize the sparse patterns of the proposed triple sparse convolution in semi-supervised VOS, to better demonstrate why the proposed method can work well.

\subsubsection{Ablation of Sparse Strategies}

As the annotated frames provide more reliable memories than unannotated frames, besides the fully sparse and mixed sparse processing strategies mentioned earlier, we also consider their intermediate which is a semi-mixed sparse processing that only applies non-sparse processing to the annotated frames. The semi-mixed sparse processing is designed to verify the importance of processing memory frames in a non-sparse manner, without the disturbance of erroneous memories. It may also serve as an alternative to trade off the performance and efficiency. Note that the memory encoder will not be sparsified when utilizing mixed sparse processing, as the memory encoder only works for the memory frames which are all processed in a non-sparse manner. However, for the fully sparse or semi-mixed sparse processing, we also consider the influence of applying sparse convolution to the memory encoder (``w/ M") or not (``w/o M").

Table \ref{table_2} lists the comparison results. We can observe that the proposed mixed sparse processing significantly improves the performance of all sparse methods on both DAVIS and Youtube-VOS datasets, compared with the fully sparse processing. In particular, the residual sparse method benefits most from the mixed sparse processing (about 36\% accuracy improvement when compared with the fully sparse processing on DAVIS), which means that the performance of the residual sparse method largely depends on the temporal reliability, especially the correctness of the extracted memories. In contrast, the proposed triple sparse method shows stronger robustness to temporal perturbations, i.e. 76.39\% accuracy is still achieved for fully sparse processing. Although the static sparse method also shows robustness to temporal perturbations, it benefits less accuracy improvement (about 2\%) from the mixed sparse processing than the triple sparse one, probably because the triple sparse method can selectively utilize the reliable temporal information while discarding those detrimental ones.

\begin{table}[!t]
\caption{Ablation study of different sparse strategies w.r.t. different sparse methods, where ``w/ M" means applying sparse convolution to the memory encoder and ``w/o M" implies not doing so. We report the $\mathcal{J}\&\mathcal{F}$ score on the DAVIS-2017 validation set (DAVIS), the overall score on the Youtube-VOS validation set (Youtube), the size of parameters (Params) and floating-point operations (FLOPs).}\label{table_2}
\begin{center}
\setlength{\tabcolsep}{3.8pt}
\renewcommand{\arraystretch}{1.2}
\begin{tabular}{cccccc}
\toprule[0.5pt]
Methods                   & Strategies & DAVIS   & Youtube & Params & FLOPs   \\ \midrule[0.5pt]
\multirow{5}{*}{Static}   & Fully (w/ M)      & 60.58\% & 60.09\% & 208.15M & 92.32G  \\
                          & Fully (w/o M)     & 73.19\% & 72.96\% & 208.00M & 103.60G  \\
                          & Semi-mixed (w/ M)  & 70.42\% & 67.70\% & 208.15M & 93.28G  \\
                          & Semi-mixed (w/o M) & 74.33\% & 74.02\% & 208.00M & 104.17G  \\
                          & Mixed             & 77.66\% & 76.39\% & 208.00M & 126.24G \\ \midrule[0.5pt]
\multirow{5}{*}{Residual} & Fully (w/ M)           & 36.42\% & 32.23\% & 208.15M & 89.30G  \\
                          & Fully (w/o M)           & 41.67\% & 42.74\% & 208.00M & 98.76G  \\
                          & Semi-mixed (w/ M)       & 26.49\% & 39.79\% & 208.15M & 101.19G \\
                          & Semi-mixed (w/o M)       & 42.98\% & 47.72\% & 208.00M & 111.21G \\
                          & Mixed             & 78.08\% & 74.11\% & 208.00M & 125.23G \\ \midrule[0.5pt]
\multirow{5}{*}{Triple}   & Fully (w/ M)            & 72.13\% & 71.17\% & 209.97M & 93.61G  \\
                          & Fully (w/o M)            & 76.39\% & 74.01\% & 209.06M & 101.43G  \\
                          & Semi-mixed (w/ M)       & 75.15\% & 74.49\% & 209.97M & 97.04G  \\
                          & Semi-mixed (w/o M)       & 79.25\% & 77.95\% & 209.06M & 107.36G  \\
                          & Mixed             & 83.04\% & 79.29\% & 209.06M & 128.42G \\ 
                          \bottomrule[0.5pt]
\end{tabular}
\end{center}
\end{table}

The semi-mixed strategy helps to achieve a better trade-off between accuracy and computation costs. For example, the triple sparse method with semi-mixed sparse processing can obtain an accuracy gain of about 3\% at the cost of about 5.8\% more FLOPs, compared with the fully sparse processing. Besides, the sparsification of the memory encoder can be regarded as an additional choice to reduce the overall FLOPs while sacrificing performance. On the basis of the proposed triple sparse method, the sparsification of the memory encoder provides about 9.6\% and 7.7\% FLOPs reduction for the semi-mixed and fully sparse processing respectively, while sacrificing about 4\% accuracy. 

It is worth mentioning that the semi-mixed sparse processing may not work well for the residual sparse method when sparsifying the memory encoder. In such case, the semi-mixed sparse processing suffers from about a 10\% accuracy drop compared with the fully sparse one on DAVIS, but achieves about 7\% higher accuracy on Youtube-VOS. There are two possible reasons: 1) The residual sparse model may not be appropriately trained, due to the overwhelming temporally accumulated errors, and consequently the evaluation results are unreliable. 2) The typical structure of the residual sparse model with sparsified memory encoder may be more prone to overfitting on the DAVIS dataset, which is much smaller than the Youtube-VOS dataset. As the residual sparse method is not our focus in this work and we do not explore more on it.

\begin{table}[!t]
\caption{Ablation study of different sparse structures w.r.t. different sparse methods, where ``Q" means the query encoder and ``D" means the decoder. "$^\dag$" means that the sparse convolution is applied for all convolution layers (including the downsampling convolution layers for skip connections). A check mark will be used to indicate that the corresponding module is sparsified. We report the $\mathcal{J}\&\mathcal{F}$ score on the DAVIS-2017 validation set (DAVIS), the overall score on the Youtube-VOS validation set (Youtube), the size of parameters (Params) 
and floating-point operations (FLOPs).}
\label{table_3}
\begin{center}
\renewcommand{\arraystretch}{1.2}
\begin{tabular}{ccccccc}
\toprule[0.5pt]
Methods                   & Q                         & D                         & DAVIS   & Youtube & Params  & FLOPs   \\ \midrule[0.5pt]
\multirow{3}{*}{Static}   & \checkmark &                           & 81.34\% & 78.49\% & 207.90M & 203.47G \\
                          &                           & \checkmark & 81.89\% & 79.49\% & 207.88M & 143.53G \\
                          & \checkmark & \checkmark & 77.66\% & 76.39\% & 208.00M & 126.24G \\ \midrule[0.5pt]
\multirow{3}{*}{Residual} & \checkmark &                           & 80.24\% & 76.50\% & 207.90M & 203.75G \\
                          &                           & \checkmark & 80.45\% & 75.04\% & 207.88M & 143.51G \\
                          & \checkmark & \checkmark & 78.08\% & 74.11\% & 208.00M & 125.23G \\ \midrule[0.5pt]
\multirow{3}{*}{Triple}   & \checkmark &                           & 82.27\% & 78.34\% & 208.48M & 204.12G \\
                          &                           & \checkmark & 84.02\% & 79.80\% & 208.36M & 143.18G \\
                          & \checkmark & \checkmark & 83.04\% & 79.29\% & 209.06M & 128.42G \\ 
                         \midrule[0.5pt]
\multirow{3}{*}{Triple$^\dag$}   & \checkmark &                           & 60.63\% & 63.51\% & 208.92M & 195.01G \\
                          &                           & \checkmark & 71.41\% & 75.46\% & 208.89M & 102.96G \\
                          & \checkmark & \checkmark & 53.65\% & 62.46\% & 210.03M & 75.66G \\ 
                         \midrule[0.5pt]
\multirow{1}{*}{Non-sparse}   &  &        & 82.88\% & 80.36\% & 207.78M & 220.62G \\
                         \bottomrule[0.5pt]
\end{tabular}
\end{center}
\end{table}

\subsubsection{Ablation of Sparse Structures}

In order to investigate the effect of the sparsification of different modules on the accuracy and computation costs, we attempt to separately sparsify different modules including the query encoder and decoder (denoted as ``Q" and ``D" respectively) in the memory-matching-based Semi-VOS framework \footnote{We have already studied the effect of the memory encoder as can be seen from Table \ref{table_2} (the  `w/ M" configurations). Thus, we do not consider ablating the memory encoder here.}. 

\begin{figure*}[!t]
\centering
\includegraphics[width=7.1in]{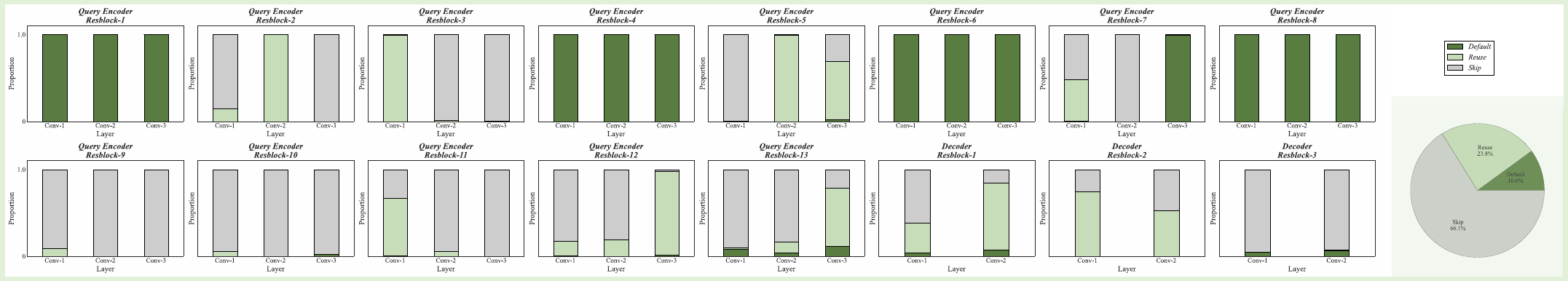}
\caption{The layer-wise sparse pattern of the proposed triple sparse convolution. Some blocks tend to be more important to extract features and thus require nearly non-sparse convolution, such as ``Resblock-1" in the query encoder. Some blocks may have more redundant information and fit better with the sparse convolution, such as the residual blocks in the decoder.}
\label{fig_layer}
\end{figure*}

As shown in Table \ref{table_3}, when separately applying sparse convolution to the query encoder or the decoder, there is no much difference among the three sparse methods except that the residual sparse method achieves slightly inferior performance on Youtube-VOS. This is because the degree of sparsity for these ablative settings is not high enough, resulting in the overall performance of the model being closer to that of the non-sparse one. On the other hand, the triple sparse method can maintain as high accuracy as possible when sparsifying both modules. For example, when sparsifying the decoder individually, the triple sparse method sacrifices only 1\% $\mathcal{J}\&\mathcal{F}$ score on the DAVIS-2017 validation set, while the static and residual sparse methods sacrifice 4.3\% and 2.4\%, respectively. One possible reason is that the proposed triple gate renders the triple sparse model more freedom, i.e. alternative three policies for each pixel, to fit more complex sparse structures.

Moreover, it can be observed from Table \ref{table_3} that the sparsification of the decoder is much more effective than the sparsification of the query encoder. Especially when comparing the triple sparse method with the non-sparse one, the sparsification of the decoder provides about 35\% FLOPs reduction with only a 0.56\% accuracy drop on Youtube-VOS. In contrast, the sparsification of the query encoder trades 2.02\% accuracy for merely 7.5\% FLOPs reduction. There may be two reasons for this: 1) The decoder takes up most of the computation in the memory-matching-based Semi-VOS framework, which in turn introduces a more significant reduction of FLOPs when sparsified. As shown in Table \ref{table_4}, the decoder accounts for approximately 75\% of the overall computation of the non-sparse method, and the computation reduction of the triple sparse method mainly comes from the decoder (approximately 87\% of the overall reduction). 2) The decoder has more feature redundancy compared to the query encoder, and thus reducing such redundancy in the decoder may lead to better performance. Generally in the memory-matching-based Semi-VOS framework, the decoder is required to refine the mask of each object and tends to focus only on the area near the objects. Such object-specific focus behavior may alleviate the disturbance of the background and thus improve performance. However, the query encoder is not object-specific and may need to consider the contextual information in the whole video frame. Consequently, the loss of contextual information after sparsification may have a greater negative impact on accuracy. A specific visualization of the above reasons will be illustrated in the next section.

\begin{figure*}[!t]
\centering
\includegraphics[width=5.8in]{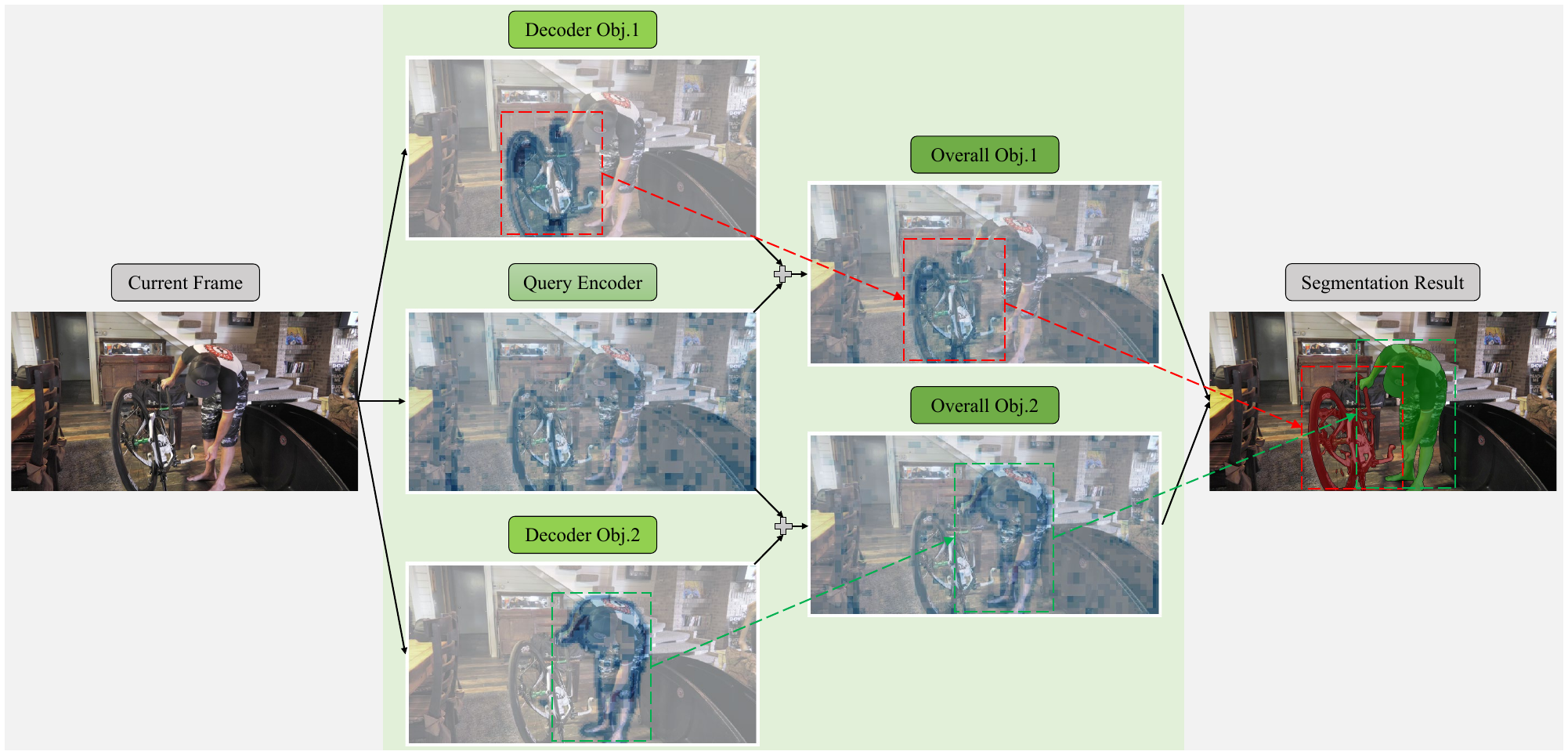}
\caption{The spatial sparse pattern of the proposed triple sparse convolution. The sparsified decoder tends to focus on specific objects. As a complement, the query encoder tends to preserve some indispensable contextual information, such as the features of the background, while being as sparse as possible.}
\label{fig_space}
\end{figure*}

\begin{table}[!t]
\caption{The detailed FLOPs of each module w.r.t. the non-sparse and triple sparse methods. FLOPs (Q), FLOPs (M), and FLOPs (D) denote the FLOPs of the query encoder, the memory encoder, and the decoder respectively.}
\label{table_4}
\begin{center}
\renewcommand{\arraystretch}{1.2}
\begin{tabular}{ccccc}
\toprule[0.5pt]
Method     & FLOPs (Q) & FLOPs (M) & FLOPs (D) & FLOPs   \\ \midrule[0.5pt]
Non-sparse & 36.00G    & 17.46G    & 167.16G   & 220.62G \\ \midrule[0.5pt]
Triple     & 24.13G    & 17.46G    & 86.83G    & 128.42G \\ \bottomrule[0.5pt]
\end{tabular}
\end{center}
\end{table}

\begin{table}[!t]
\caption{The ablation of the temperature in Gumbel-Softmax. The ``Accuracy" denotes the $\mathcal{J}\&\mathcal{F}$ score on the DAVIS-2017 validation set. The ``FLOPs" is floating-point operations. The sparsity target is fixed at 0.1 for all temperatures.}
\label{table_5}
\begin{center}
\setlength{\tabcolsep}{3.3pt}
\renewcommand{\arraystretch}{1.2}
\begin{tabular}{cccccccc}
\toprule[0.5pt]
Temperature & 0.01    & 0.1     & 0.5     & 1       & 5       & 10      & 100     \\ \midrule[0.5pt]
Accuracy (\%)   & 78.28 & 81.32 & 82.40 & 83.04 & 83.43 & 82.43 & 82.79 \\ \midrule[0.5pt]
FLOPs (G)      & 126.91 & 128.34 & 126.20 & 128.42 & 136.09 & 132.80 & 127.06 \\ \bottomrule[0.5pt]
\end{tabular}
\end{center}
\end{table}

It is worth mentioning that applying sparse convolution to all convolution layers will introduce severe performance degradation, especially for the query encoder (22.25\% accuracy drop with only 11.6\% FLOPs reduction when compared with the non-sparse method on DAVIS). This may be due to the overly sparse structure, which sparsifies the downsampling convolution layers that already compress information and thus results in more severe information loss and performance degradation. Similar to the previous discussions, the decoder exhibits better fitness for the sparsification of all convolution layers (over 50\% FLOPs reduction). However, the model still suffers a severe accuracy drop of about 10\% on DAVIS due to excessive loss of information. Note that the most sparse case, which applies sparse convolution to all convolution layers in the query encoder and the decoder, will provide the most significant reduction in FLOPs of about 66\%. However, in this case, the model will no longer be able to maintain acceptable performance.

\subsubsection{Ablation of Gumbel-Softmax Temperature}

To study the effect of the temperature in Gumbel-Softmax, we explore a variety set of temperatures from 0.01 to 100, and the results are shown in Table \ref{table_5}. The FLOPs of different temperatures do not differ much from each other (deviations are about -1.7\%$\sim$3.4\% w.r.t. the FLOPs of temperature 1), indicating that the sparsity is temperature-irrelevant. For temperatures higher than 1, no significant accuracy variation (1\% larger or lower than the accuracy where temperature equals 1) is observed. However, for temperatures much lower than 1, such as 0.1 and 0.01, the accuracy drops drastically. One possible explanation is that the lower the temperature, the more the Softmax operation tends to be the Argmax operation w.r.t. the probabilities of policies. Small noise on the probability may lead to an absolutely erroneous one-hot result after Gumbel-Softmax, biasing the triple gate to generate a suboptimal policy. Considering the Gumbel noise term ($\boldsymbol{G}$) in Eq. \ref{eq_9}, random noise is consistently imposed to the log-softmaxed probabilities ($\log \left( \text{Softmax} \left( \boldsymbol{P} \right) \right)$). With a low temperature, the imposed noise may deviate the probabilities and misguide the triple gate to generate an improper policy, especially in the initial phase of training where the triple gate is not well trained to generate robust policies. Consequently, ill-trained triple gates will lead to inferior accuracy.

\subsubsection{Sparse Patterns in Semi-VOS}

In order to better understand how the proposed sparse convolution helps the Semi-VOS task, we give an in-depth analysis of the produced sparse patterns in different layers and at different feature positions of the Semi-VOS framework. This further inspires the design of the sparse convolutional modules.

Fig. \ref{fig_layer} shows the layer-wise sparse pattern produced by the proposed triple sparse convolution in Semi-VOS. An interesting observation is that the ``Resblock-1", ``Resblock-4", ``Resblock-6" and ``Resblock-8" in the query encoder require almost non-sparse processing, i.e. 100\% default policies. At the same time, other blocks mostly rely on the information from the previous frame with the skipping filter. This reflects the inherent sparsity property residing in the semi-supervised VOS task, that is, some residual blocks such as ``Resblock-1" in the query encoder tend to capture more essential features of the object and should not be sparsified, while other blocks may contain redundant information similar to that in the previous frame and can be simply omitted. Based on this, the proposed SpVOS has good acceleration potential in a fine-grained layer-wise manner, by discarding those less important convolution layers while applying non-sparse convolution to the important ones. In addition, the percentages of the three policies over all layers are shown in the pie chart at the bottom right of Fig \ref{fig_layer}, where the skipping policies account for the largest part of 66.1\% followed by the reusing policies with a ratio of 23.8\%. This means that there is more redundant information (about 1.8 times more) that can be skipped or filtered compared to reusable temporal information in Semi-VOS.  

Besides the layer-wise sparse pattern, we also explore the spatial sparse pattern produced by the proposed triple sparse convolution, which is obtained by accumulating the number of convolutions performed at each spatial location and can be seen in Fig. \ref{fig_space}. It can be observed that the decoder tends to become object-specific after being sparsified by only applying convolutions around the object-of-interest, while the query encoder seems to be object-agnostic and attempts to capture global information from the whole video frame, which accords with our discussion in the previous section. As a result, the overall spatial sparse patterns focus more on the object-of-interest due to the object-specific sparse decoder, but still maintain the ability to capture the context information from the background owing to the object-agnostic sparse query encoder.

\subsubsection{Adaption to SOTA}

Instead of directly comparing with SOTA methods, which are difficult due to heterogeneous frameworks. We attempt to demonstrate the immense potential of the proposed triple sparse convolution via adapting it to a SOTA framework named XMem \cite{cheng2022xmem}. The training configurations follow the ones of SpVOS except for a minor difference: Four frames are sampled in a batch to form a video clip, within which two frames (the first frame and a random frame chosen from the second and third ones) are selected as memory frames. According to results in Table \ref{table_6}\footnote{The results of XMem and ``Tri-XMem" in Table \ref{table_6} are the ones without pertaining on the BL30K dataset.}, when combined with the triple sparse convolution (``Tri-XMem"), the accuracy of XMem does not significantly deteriorate, even achieving 1.4\% improvement on the DAVIS-2017 validation set. With negligible 0.71\% parameter overhead, 34.56\% FLOPs reduction and $1.26\times$ real-time acceleration (for V100 GPU) are obtained. The results imply the possibility of combining the proposed triple sparse convolution with SOTA frameworks to reduce their computation overhead without sacrificing performance. Adapting the triple sparse convolution to sparsify other SOTA frameworks would be a possible future direction.

\begin{table}[!t]
\caption{Adapting the proposed triple sparse convolution to a SOTA framework, named XMem \cite{cheng2022xmem}. We report the $\mathcal{J}\&\mathcal{F}$ score on the DAVIS-2017 validation set (DAVIS), the overall score on the Youtube-VOS validation set (Youtube), the size of parameters (Params), floating-point operations (FLOPs), and frames per second (FPS).}
\label{table_6}
\begin{center}
\setlength{\tabcolsep}{7pt}
\renewcommand{\arraystretch}{1.2}
\begin{tabular}{cccccc}
\toprule[0.5pt]
Methods  & DAVIS  & Youtube & Params  & FLOPs   & FPS   \\ \midrule[0.5pt]
XMem \cite{cheng2022xmem}   & 86.20\% & 85.70\%  & 237.22M & 311.95G & 25.46 \\ \midrule[0.5pt]
Tri-XMem & 87.60\% & 85.50\%  & 238.91M & 204.15G & 32.15 \\ \bottomrule[0.5pt]
\end{tabular}
\end{center}
\end{table}

\section{Conclusion}

This paper explores to design an efficient semi-supervised video object segmentation pipeline by introducing the triple sparse convolution coupled with mixed sparse processing. It achieves a significant reduction of computation costs while still maintaining a high segmentation accuracy. Compared with existing sparse convolution methods, the triple sparse convolution can better handle the segmentation error accumulation and similar object confusion due to the proposed triple gate, which utilizes temporal information to help distinguish similar objects and adaptively filters detrimental spatial information to alleviate the propagation of errors. Experimental results on two benchmark VOS datasets well validate the effectiveness of the proposed triple sparse convolution, and we hope that SpVOS can serve as a baseline for future sparse VOS researches and inspires more efficient sparse VOS works.


%



\section*{Acknowledgment}
This work is supported by National Natural Science Foundation of China (No. 62071127),  National Key Research and Development Program of China (No. 2022ZD0160100), Shanghai Natural Science Foundation (No. 23ZR1402900).


\ifCLASSOPTIONcaptionsoff
  \newpage
\fi



\bibliographystyle{IEEEtran}
\bibliography{IEEEabrv,./refs}
\end{document}